\documentclass[twoside,11pt]{article}
\usepackage{pgm}

\usepackage[utf8]{inputenc}
\inputencoding{utf8}
\usepackage{times}

\usepackage[para]{footmisc}
\usepackage{soul}
\usepackage{url}
\usepackage{subcaption}
\usepackage{graphicx}
\usepackage{amsmath}
\usepackage{booktabs}
\usepackage{pgf,tikz}
\usepackage{url}
\usepackage{algorithmicx}
\usepackage{algorithm}
\usepackage[noend]{algpseudocode}
\usepackage{multicol}
\usepackage{booktabs}
\usepackage{xcolor}
\usepackage{colortbl}
\usepackage{multirow}
\usepackage{stackrel}

\usepackage{mathtools} 
\usepackage{booktabs} 
\usepackage{tikz} 

\usetikzlibrary{arrows,automata}
\newcommand{\graph}{\mathit{G}}
\newcommand{\score}[2]{\sigma_{#1}({#2})}
\newcommand{\vertex}[1]{V_{#1}}

\newcommand{\edges}{\mathit{E}}

\newcommand{\parents}{\Pi}

\usepackage{xspace}
\newcommand{\problem}{\text{BNSL\xspace}\xspace}

\newcommand{\hrulealg}[0]{\vspace{1mm} \hrule \vspace{1mm}}
\newcommand{\Statey}{\Statex\hspace*{\ALG@thistlm}}
\title{Scalable Bayesian Network Structure Learning with Splines}
\usepackage{hyperref,color,soul,booktabs}
\setulcolor{blue}
\newcommand{\BibTeX}{\textsc{B\kern-0.1emi\kern-0.017emb}\kern-0.15em\TeX}

\DeclareMathOperator*{\argmin}{argmin}

\begin{document}


 \author{\Name{Charupriya Sharma} \Email{c9sharma@uwaterloo.ca}\and
   \Name{Peter van Beek} \Email{vanbeek@uwaterloo.ca}\\
   \addr David R. Cheriton School of Computer Science, University of Waterloo}

\maketitle

\begin{abstract}
The graph structure of a Bayesian network (BN) can be learned from data using the well-known score-and-search approach. Previous work has shown that incorporating structured representations of the conditional probability distributions (CPDs) into the score-and-search approach can improve the accuracy of the learned graph. In this paper, we present a novel approach capable of learning the graph of a BN and simultaneously modelling linear and non-linear local probabilistic relationships between variables. We achieve this by a combination of feature selection to reduce the search space for local relationships and extending the score-and-search approach to incorporate modelling the CPDs over variables as Multivariate Adaptive Regression Splines (MARS). MARS are polynomial regression models represented as piecewise spline functions. We show on a set of discrete and continuous benchmark instances that our proposed approach can improve the accuracy of the learned graph while scaling to instances with a large number of variables.
\end{abstract}

\section{Introduction}
Bayesian networks (BNs) are widely used probabilistic graphical models with applications in knowledge discovery, decision support, and prediction \citep{Darwiche09, KollerF09}. 
A BN represents a joint distribution over random variables as a product of conditional probability distributions (CPDs) for each variable, and it enforces a directed acyclic graph (DAG) structure over the corresponding network. The \emph{score-and-search} approach is often used to learn the structure of a BN from data, where a \emph{scoring function} is used to evaluate the quality of fit of a BN over the data. 

The Bayesian Information Criterion (BIC) \citep{schwarz1978estimating} is one of the most widely used scoring functions to learn BNs. It is expressed as the difference of a likelihood and a penalty term for model complexity, and gives theoretical guarantees that enable significant pruning of the search space of DAGs. However, in practice, BIC tends to heavily penalise large parent sets in a DAG, making it less accurate for datasets which need more complex models. 

Implementations of score-and-search usually employ simple conditional probability tables (CPTs) to represent the CPD for a variable (e.g., \citep{BartlettC13, Chickering2002, vanBeek2015}). In settings with limited data or with a large number of variables, such a representation might not be efficient, because CPTs grow exponentially with the size of the parent set. 
Using CPTs can also result in overfitting. These two disadvantages have led to research of structured representations for CPDs that can model complex relationships between a child and its parents without exponential size and overfitting risks.

One example of a structured representation is context-specific independence \citep{boutilier96}, which is a popular method to represent independence given a certain assignment of a subset of variables. Tree structures were extended to be represented as decision graphs \citep{chickering1997bayesian} with equality constraints. However, the partitioning algorithms used to split the feature space become computationally infeasible for discrete variables with large domains. Other frequently used representations include the Noisy-OR \citep{Pearl1988}, which assumes a causal independence within the parent set and allows CPTs to be represented with parameters linear in the size of the parent set, and logistic regression \citep{buntine1991theory}. Most studies of structured representations e.g., \citep{xiang2007modeling, zhang96exploiting} do not impose global structural constraints on the DAG.

\cite{friedman1998learning}~were the first to extend the \emph{score-and-search} approach to include  structured representations in Bayesian network structure learning (BNSL). They showed this improved results in inference and allowed exploration of more complex network structures. \cite{talvitie2019learning} gave a tree-based score to find the optimal BN with some structural constraints. Their experiments with structured CPDs showed an improvement in the ability of the search algorithm to find correct BNs with less data on some real-world datasets, but on other datasets, CPTs performed better. \cite{sharma2020score}  extended score-and-search to the Noisy-OR representation and let the search algorithm have the choice between a structured and a Noisy-OR representation. They observed their learned networks to have a mixture of Noisy-ORs and standard CPTs, which improved overall network scores. This shows that there needs to be some flexibility about the structured representation used for the CPD to enable choosing the best fitting representation to find the correct overall structure and maintain the complexity advantage in inference. 

Other approaches to scale beyond small network sizes include using neural networks to solve a continuous formulation of the DAG learning problem \citep{ yu2019dag, zheng2018dags}.
\citet{scanagatta2015learning} used approximate search for discrete  benchmarks, but did not allow continuous variables or structured representations.

In this paper, 
 we propose the first score-and-search approach for learning Bayesian networks with  Multivariate Adaptive Regression Splines (MARS) as possible representations for CPDs and do not make a priori assumptions on the structure of the DAG. 
MARS are polynomial regression models represented as piecewise spline functions.
 Our approach can effectively prune most candidate parent sets of a variable by leveraging feature learning algorithms. 

Our contributions are the following:
\begin{itemize}
    \item Our scoring method can model both linear and non-linear interactions between variables without imposing any rigid constraints on their structure while scoring them simultaneously (Section~\ref{SECTION:MARS}). Variables are scored independently of each other.
    \item Our scoring algorithm can identify high-degree variable interactions without a limit on the number of variables or terms in the model (Section~\ref{SECTION:MARS}).
    \item Our scoring algorithm can compute structured representations for both discrete and continuous valued variables (Section~\ref{SECTION:MARS}).
    \item We provide a feature selection method to identify the most promising parents for a node, and scale our score-and-search approach to very large networks with 1000+ nodes (Section~\ref{SECTION:FeatureSelection}).
    \item
    We show empirically on a set of discrete and continuous benchmark instances that our proposed approach can improve the accuracy of the learned DAG (Section~\ref{SECTION:Experimental}).
\end{itemize}

\section{Preliminaries}

		
		
		

 A BN represents a joint distribution  over random variables  $\mathbf{V} = \{V_{1}, \ldots, V_{n}\}$ with a labeled directed acyclic graph (DAG), $\graph = (\mathbf{V}, \edges)$ where the edges $\edges$ represent direct influence of one random variable on another, and each node $\vertex{i}$ is labeled with a conditional probability distribution $P(V_{i} \mid \Pi_{i})$ that defines the relationship of the variable $V_{i}$ with its set of parents $\Pi_i$ in the DAG and $P(V_{1}, \ldots, V_{n}) = \prod_{i=1}^n P(V_i \mid \Pi_i).$ Each variable $V_i$ has a state space $\Omega_i$. Each set of parents $\parents_{i} = \{V_{i1}, \ldots, V_{i|\parents_{i}|}\}$ has as state space the set of candidate instantiations of the nodes in $\parents_{i}$, $\Omega_{\parents_{i}} = \Omega_{i1} \times \dots \times \Omega_{i|\parents_{i}|}$.

\textbf{Score-and-search:} Given a set of $N$ instances $I = \{I_1, \ldots, I_N\}$, where each instance $I_i$ is an $n$-tuple that is a complete instantiation of the $n$ variables in $\mathbf{V}$, a \emph{scoring function} $\sigma( \graph \mid I )$ assigns a real value to a DAG $\graph$ to measure the quality of its fit to data $I$. When the dataset is clear from context, we express this as  $\sigma(\graph)$. The objective of Bayesian network structure learning (\problem)  is to find the network $\argmin_{G} \sigma(G)$, and a lower score is assumed to represent a better quality fit without loss of generality.



\section{MARS: Multivariate Adaptive Regression Splines}
\label{SECTION:MARS}

Linear models like linear and logistic regression are a simple, intuitive and fast to compute way to represent data. However, they impose linear relations on the data which limits their accuracy if the ground truth contains non-linear relationships. Multivariate Adaptive Regression Splines (MARS) \citep{friedman1991multivariate} are an extension of linear models that combines linear basis functions to model non-linearities and variable interactions.
MARS models are of the form,
\begin{equation}
 M(x) = c_0 + \sum_{j=1}^{k}c_{j}B_{j}(x) ,
\label{eq:mars_model}
\end{equation}
where each \emph{model coefficient} $c_j$ for $j \in [0,k]$ is a constant, and each $B_j$ is a \emph{basis function}. Basis functions take one of the two forms:
\begin{enumerate}
 \item A \emph{hinge} function of the form $(x - t)_+ = \max(0, x-t)
   \text{ or }\ (t - x)_+ = \max(0, t-x) $, where $t$ is a constant, called a \emph{knot}. 
 \item A product of two or more unique hinge functions. This prevents terms of higher degree which are approximated by the hinge functions instead.
\end{enumerate}

\begin{figure}[thb]
  \begin{minipage}[c]{0.50\textwidth}
    \caption{
       MARS (red) and linear regression (blue) models for the same set of points. MARS places a knot at every change of the slope.
    }\label{FIGURE:MARS}
  \end{minipage}
  \begin{minipage}[c]{0.45\textwidth}
    \centerline{\fbox{\includegraphics[scale=0.6]{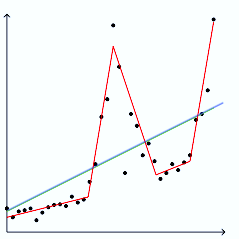}}}
  \end{minipage}\hfill
\end{figure}

Hinge functions are zero over part of their range, and their product is nonzero only over the region where each of the component hinge functions have nonzero values. Thus, data can be partitioned into disjoint regions and processed independently, enabling processing of high dimensional inputs while still being able to capture complex non-linear relationships with high order interactions (see Figure~\ref{FIGURE:MARS}).

\textbf{Variable modelling:} 
For a variable $V_i$ and a candidate parent set $\Pi_i$ we write the MARS model for $V_i$ and $\Pi_i$ as,
\begin{equation}
 M(\Pi_i) := c_0 + \sum_{j=1}^{k}c_{j}B_{j}(\Pi_i) .
\label{eq:mars_local}
\end{equation}
For the computation of the model, we are given a set of $N$ instantiations $I = \{I_1, \dots, I_N\}$ for $V_i$ and $\Pi_i$, i.e. $I_j = (y_i,x_1,\dots,x_{|\Pi_i|})$ with $y_i \in \Omega_i, x_j \in \Omega_{ij}$. 

\textbf{Model computation:} To compute the basis functions in $M(\Pi_i)$, the MARS algorithm operates in two phases: 1) A forward stage which starts with the intercept $c_0$ and proceeds to generate many candidate basis functions in pairs $[(t - x)_+,  (x - t)_+]$ with $x \in \Pi_i$. The model coefficients are estimated by minimizing the residual sum-of-squares (RSS),
\begin{equation}
    RSS(M(\Pi_i)) = \sum_{j=1}^N \left(y_j - M(I_j)\right)^2  .
    \label{eq:rss}
\end{equation}
A candidate pair of basis functions is added to the model $M(\Pi_i)$ if it reduces the training error of $M(\Pi_i)$; i.e.~if $RSS(M_{new}(\Pi_i)) - RSS(M_{old}(\Pi_i)) < 0$. This process continues until there is no significant improvement. 2) A backward stage that is used to address overfitting in $M(\Pi_i)$. The backward stage iterates over functions in $M(\Pi_i)$ and prunes terms until the subset of the terms that provide the best score remains. The subsets are scored by using a generalized cross-validation (GCV) score,
\begin{equation}
    GCV(M(\Pi_i)) = \frac{RSS(M(\Pi_i))}{((1 - \lambda)/N)^2} ,
    \label{eq:gcv}
\end{equation}
where $\lambda$ is the number of terms in $M(\Pi_i)$. We call $GCV(M(\Pi_i))$ the local score of $V_i$ in a BN which assigns $\Pi_i$ as the parent set of $V_i$. GCV provides a computationally fast approximation to leave-one out cross-validation, for linear fitting under squared-error loss. The GCV score can be shown to be similar to the Akaike information criterion (AIC) score, which is frequently used for model selection \citep{hastie2009elements}. 

\textbf{Scoring BNs:} GCV scores are decomposable, meaning for a BN $G$ with parent set $\Pi_1, \dots, \Pi_i$ assigned to variables $V_1,\dots, V_n$, the total GCV score can be calculated by,
\begin{equation}
    GCV(G) = \sum_{i=1}^n GCV(M(\Pi_i)),
\end{equation}
where $M(\Pi_i)$ is the model computed for representing $V_i$ with candidate parent set $\Pi_i$.

\section{Scaling with Feature Selection}\label{SECTION:FeatureSelection}
\tikzstyle{decision} = [diamond, draw, fill=blue!20, 
    text width=4.5em, text badly centered, inner sep=0pt]
\tikzstyle{block1} = [rectangle, draw, fill=green!20]
\tikzstyle{block} = [rectangle, draw, fill=blue!20,  text centered, rounded corners]
\tikzstyle{block2} = [rectangle, draw, fill=red!20,  text centered, rounded corners]
\tikzstyle{line} = [draw, -latex']
\tikzstyle{cloud} = [draw, ellipse,fill=red!20]
    


MARS provides a fast way to score a candidate parent set for a child. However, there are an exponential number of candidate parent sets for each node. To scale the search to large networks, it is crucial to score a small subset of candidate parent sets. In this section, we describe our algorithm BN-FS-MARS  (Algorithm~\ref{alg:BNMARS}) for $\problem$ which is based on MARS and a feature selection mechanism. 

BN-FS-MARS operates in two phases: 1) Computing candidate MARS models and their GCV scores for all variables and candidate parent sets. We prune candidate parents for a variable $V_i$ by using feature selection: we compute an ordering for the set of candidate parents $\mathbf{V} \setminus\{V_i\}$ which ranks their importance, and use the variable importance rankings to create candidate parent sets.
2) The first phase provides multiple possible parent sets for each variable with an associated score for each parent set and this gives us an optimisation problem of finding the DAG $G$ that minimizes the total score. We feed the candidate parent sets and scores from the MARS models of phase 1 into a search algorithm to construct a DAG of minimum GCV score. Algorithm~\ref{alg:BNMARS} 
shows the pseudocode of the algorithm. 


The performance of BN-FS-MARS has significant dependence on the quality of the variable importance ranking used in Step 3 of Algorithm \ref{alg:BNMARS}. There are a number of feature selection methods such as random forests \citep{breiman2001random}, decision trees, and the Pearson correlation coefficient that can generate such a ranking (see, e.g., \citep{guyon2003introduction} and references therein). The accuracy of the feature selection method used to generate this ranking depends significantly on the amount and type of data we have available.
In the next section, we will show the performance of BN-FS-MARS with random forests.

\begin{algorithm}[htb]
\caption{BN-FS-MARS\label{alg:BNMARS}}
\begin{algorithmic}[1]
\Statex \textbf{Inputs}: A set of random variables $\mathbf{V}$ and a set of observations $I$
\Statex \textbf{Outputs}: A DAG $G$ minimizing GCV($G$)
\hrulealg
\Statex \emph{Phase 1: Compute MARS models and GCV scores for all variables guided by random forests}
\State Initialize $R = \emptyset$ to collect scores for Phase 2
\For {each variable $V_i \in \mathbf{V}$}
    \State $\mathbf{\Pi_i} :=  \texttt{FilterCandidatesFS($V_i$)}$
    \For {$k = 0,\dots,|\mathbf{\Pi}_i|$}
        \For {parent sets $\Pi_i \subseteq \mathbf{\Pi}_i, |\mathbf{\Pi}_i| = k$}
            \State Compute MARS model $M(\Pi_i)$ \hfill \emph{see Eqn.~(\ref{eq:mars_local})}
            \State Compute $GCV(M(\Pi_i))$ \hfill \emph{see Eqn.~(\ref{eq:gcv})}
            \State $R = R \cup \{(V_i,\Pi_i,GCV(M(\Pi_i)))\}$
            \If {time limit for $V_i$ reached}
                \State Continue for-loop in line 2
            \EndIf
        \EndFor
    \EndFor
\EndFor
\Statex \emph{Phase 2: Choose MARS models for each variable to create a DAG that minimizes $GCV(G)$}
\State Run a search algorithm on $R$ to compute a DAG $G$ minimizing $GCV(G)$
\end{algorithmic}
\end{algorithm}


\begin{algorithm}[htb]
\caption{FilterCandidatesFS\label{alg:filterRF}}
\begin{algorithmic}[1]
\Statex \textbf{Inputs}: A random variable $V_i \in \mathbf{V}$, a constant $\lambda < n$
\Statex \textbf{Outputs}: A candidate parent set $\Pi_i$
\hrulealg
\State Compute a random forest for $V_i$ using $\mathbf{V} \setminus \{V_i\}$
\State Return $\mathbf{\Pi}_i := $ $\lambda$-most important variables based on feature selection. 
\end{algorithmic}
\end{algorithm}

\section{Experimental Evaluation}\label{SECTION:Experimental}

In this section, we show the performance of our algorithm in computing DAGs for a given dataset.  All experiments are conducted on computers with 2.2 GHz Intel E7-4850V3 CPUs with a memory limit of 32 GB, and each score-and-search experiment is limited to a total of 24 hours.

\textbf{Datasets:} We test the performance of our learned networks on synthetic networks generated using the software Tetrad \footnote{\url{https://github.com/cmu-phil/Tetrad}}. In addition, we show results on real-world Bayesian networks from the Bayesian Network Repository \footnote{\url{www.bnlearn.com/bnrepository}} for large and very large network sizes. We also show results on 100 node networks in the DREAM 4 challenge benchmarks, which are gene regulation networks using the genenetweaver software \footnote{\url{http://gnw.sourceforge.net}}. Gene regulation networks are a collection of biological regulators that interact with each other, and this interaction can be modelled as a directed graph.

\textbf{Comparison to other algorithms:}  We ran our structure learning algorithm, BN-FS-MARS, on the datasets to learn the DAG and compared the results obtained with tabular (CPT-based) BIC and Greedy Equivalent Search \citep{chickering2002optimal}, which is a score-based local search algorithm that searches over the space of equivalence classes of Bayesian-network structures. Tabular BIC results are computed using GOBNILP\citep{BartlettC13} (see below) with standard pruning rules applied, and GES results are reported using the R package pcalg (v2.7-4) \footnote{\url{https://cran.r-project.org/web/packages/pcalg/}} with the score as BIC. For tabular BIC, parent set size was limited to 2, as scoring with larger parent set sizes lead to the vast majority of the experiments failing with an out of memory (OM) error. We refer to this variant of tabular BIC as BIC2. We also experimented with the neural network based DAG-GNN, however as observed in \citep{yu2019dag}, we noted that the performance was weaker than that of GOBNILP, so in our experiments, we have reported the GOBNILP results. In addition, we experimented with WINASOBS from \cite{scanagatta2015learning}, however within our experimental setup it was outperformed by GES, and thus we have not reported the results in detail.

\textbf{Performance evaluation metrics: } We show the performance of BN-FS-MARS using the structural Hamming distance (SHD) and the F1-score. Given a ground truth network and a learned network, SHD measures the distance between the completed partially directed acyclic graphs (CPDAG) representations of the networks, where a CPDAG captures the equivalence class to which a network belongs (see~\cite{tsamardinos2006max}). It is the sum of false negatives (FN), which are missing edges; false positives (FP) which are extra edges; and edges with wrong orientations (WD). 
We use the undirected graph (the skeleton of the DAG) for computing F1-scores. Given the number of edges present (SP) in both the skeleton of the ground truth network and the skeleton of the learned network, i.e. the correct edges, the false positive (FP) and the false negatives (FP), we can compute the F1-score with $\frac{\textrm{SP}}{\textrm{SP}+\frac{1}{2}(\textrm{FP}+\textrm{FN})}$. The F1-score is between 0 and 1, with 1 being the best value.

\subsection{Implementation Details} \label{subsec:implementation}
\textbf{Computing local scores with MARS: } MARS models for representing parent-child relationships were computed using the earth package (v5.3.1) \footnote{\url{https://cran.r-project.org/web/packages/earth}}  in R. This package builds regression models using the techniques proposed by \citep{friedman1991multivariate}. The earth package limits every term in the model to have at most ten hinge functions in one product. However, for our experiments, we did not find this to be a limitation, as the models we learned did not exceed five hinge functions in a product. 

\begin{table*}[htb]
\label{tab:bnlearn}
\centering
\begin{tabular}{llrr|llrr|llrr}
id & network & $n$ & $m$ & id & network & $n$ & $m$ & id & network & $n$ & $m$ \\
\hline
0 & pathfinder & 109 & 195 & 1 & munin1  & 186  & 273  & 2 & andes   & 223 & 338  \\
3 & diabetes & 413 & 602 & 4 & pigs & 441 & 592 & 5 & link & 724 & 1125 \\
6 & munin2 & 1003 & 1244 & 7 & munin4 & 1038 & 1306 & 8 & munin3 & 1041 & 1388\\ 9 & munin  & 1041  & 1397 \\  
\hline
\end{tabular}
\caption{Benchmarks in bnlearn.}\label{TABLE:bnlearn}
\end{table*}


 To process discrete data for generating a MARS model, earth splits a categorical variable  into $l$ indicator columns of 1s and 0s, where $l$ is the number of unique values the variable can take.  For categorical response variables with $l$ values, earth computes $l$ models simultaneously. While the basis functions remain constant across these models, the coefficients can differ. Forward and backward phases are performed with the GCVs and RSSs summed across all $l$ to minimize the sum of the GCV scores across all $l$ models.
 
\textbf{Feature selection with random forests:}
Random forests are an ensemble learning method that operate by constructing a set of decision trees at training time and outputting the class that is the average prediction  of the individual trees for regression problems. Training random forests involves applying bagging to tree learners. Given a training set, bagging repeatedly selects a random sample of the training data and fits a decision tree to that sample.
Random forests can rank variables by importance in a regression problem as follows: 1) Fit a random forest to the data set and record the out-of-bag error for each observation (averaged over the trees in the forest). 2) The importance of the variable $V_i$ is then determined by perturbing the values of variable $V_i$ in the training data and computing the out-of-bag error again on the perturbed data set. The importance score for
variable $V_i$ is then determined by averaging the difference of the out-of-bag error before and after the perturbation over all trees in the random forest and normalizing it by the standard deviation of the differences.
Random forests were computed using H2O (v3.32.0.5) \footnote{\url{https://docs.h2o.ai}} in Python. 

\textbf{Finding DAGs of minimum score:} To solve the optimisation problem of finding the DAGs with the minimum total GCV score, we use GOBNILP, which is a state-of-the-art integer programming method for finding an optimal Bayesian network given a list of candidate parent sets and local scores for each variable $V_i$, with the objective of solving \problem to find a DAG $\graph$ that minimises the total score, $\score{}{G}$. GOBNILP is an anytime algorithm, and it will return the best network at a given time.

\begin{figure}[th]
\renewcommand\thesubfigure{\arabic{subfigure}}
\begin{subfigure}[b]{0.55\textwidth}
   \includegraphics[scale=0.75]{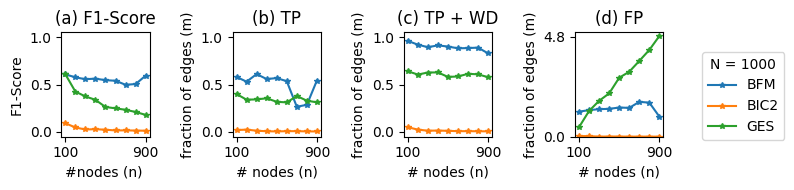}
\end{subfigure}

\begin{subfigure}[b]{0.55\textwidth}
   \includegraphics[scale=0.75]{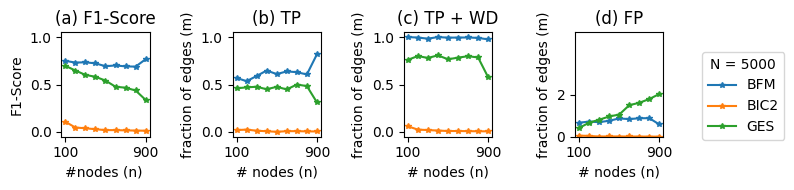}
\end{subfigure}
\caption{Comparison of structure learning approaches on various networks from Tetrad, where $n$ is the number of variables and the number of edges in the ground truth network: our BN-FS-MARS (BFM), tabular BIC with maximum parent set size as 2 (BIC2), and GES with the BIC score (GES). For each ground truth network, we report the following for corresponding learned networks for sample sizes $N=1000$ and $N=5000$: (a) F1-score, (b) TP (true positives), (c) sum of TP and WD (wrong direction), and (d) FP (false positives).
}\label{fig:Tetrad}
\end{figure}

\begin{figure}[th]
\renewcommand\thesubfigure{\arabic{subfigure}}
\begin{subfigure}[b]{0.55\textwidth}
   \includegraphics[scale=0.75]{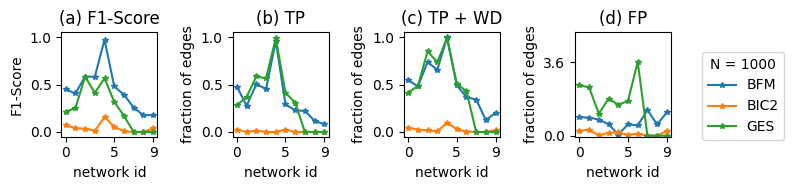}
\end{subfigure}

\begin{subfigure}[b]{0.55\textwidth}
   \includegraphics[scale=0.75]{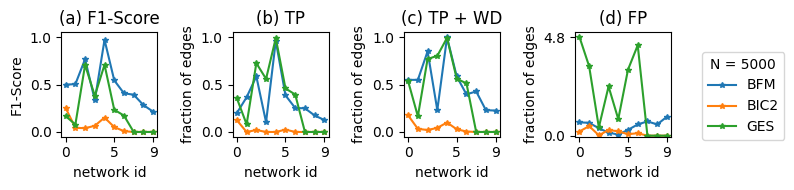}
\end{subfigure}
\caption{Comparison of structure learning approaches on various networks from bnlearn: our BN-FS-MARS (BFM), tabular BIC with maximum parent set size as 2 (BIC2), and GES with the BIC score (GES). For each ground truth network, we report the following for corresponding learned networks for sample sizes $N=1000$ and $N=5000$: (a) F1-score, (b) TP (true positives), (c) sum of TP and WD (wrong direction), and (d) FP (false positives).}\label{fig:bnlearn}
\end{figure}

\begin{figure}[th]
\renewcommand\thesubfigure{\arabic{subfigure}}
\begin{subfigure}[b]{0.55\textwidth}
   \includegraphics[scale=0.75]{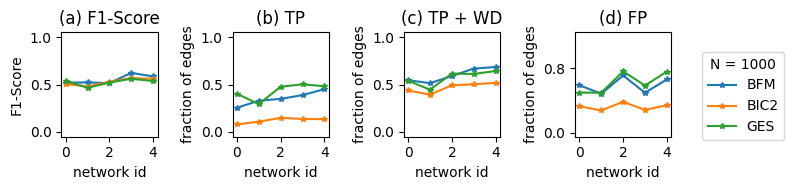}
\end{subfigure}

\begin{subfigure}[b]{0.55\textwidth}
   \includegraphics[scale=0.75]{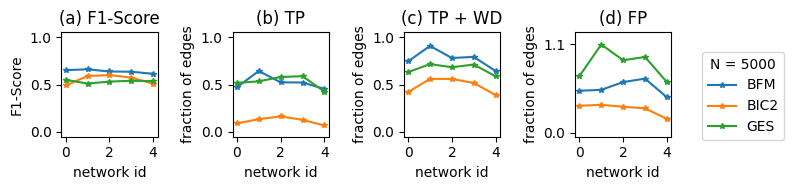}
\end{subfigure}

\caption{Comparison of structure learning approaches  on various $n=100$ networks from Dream4: our BN-FS-MARS (BFM), tabular BIC with maximum parent set size as 2 (BIC2), and GES with the BIC score (GES). For each ground truth network, we report the following for corresponding learned networks for sample sizes $N=1000$ and $N=5000$: (a) F1-score, (b) TP (true positives), (c) sum of TP and WD (wrong direction), and (d) FP (false positives).} \label{fig:dream}
\end{figure}

\subsection{Performance on Discrete Variables }

 To compare the performance of Algorithm \ref{alg:BNMARS} on discrete data, we generated synthetic networks of 100-900 nodes with Tetrad, with the maximum in-degree and out-degree of each node set as 3.  We also used real world networks from bnlearn, with sizes betwen 100-1000+ nodes (see Table~\ref{TABLE:bnlearn}).  From both these sets of networks, we randomly generated datasets with 1000 and 5000 samples, and scored candidate parent sets of all nodes with BN-FS-MARS using the top 5 features ranked by random forests in Step 3 of Algorithm \ref{alg:BNMARS}. Finally, we computed the DAG structures using GOBNILP (Step 11), and compared the performance of BN-FS-MARS with BIC2 and GES.
 Due to the anytime nature of the algorithm, we are able to obtain the best network found at any given time, and we use this property to report results on the network found within the time limit of 24 hours for score-and-search.

\textbf{Performance on F1-scores:} Figure  \ref{fig:Tetrad} reports the F1-score of the synthetic Tetrad networks learned by each of the three methods in column (a). We observe that BN-FS-MARS(BFM) has the highest F1-scores, followed by GES. 
Figure  \ref{fig:bnlearn} reports the F1-score of the bnlearn networks learned by each of the three methods in column (a). BN-FS-MARS usually has the best F1-scores, especially as $n$ increases, and the second best F1-scores are from GES. 
For both Tetrad and bnlearn, we observe that increasing sample size from $N=1000$ to $N=5000$ improves F1 scores of BN-FS-MARS and GES, and BN-FS-MARS continues to have the highest scores. However, adding more data does not improve the F1-scores for BIC2.

\textbf{Performance on SHD:} Figure \ref{fig:Tetrad} shows results for Tetrad on networks learned by each of the three methods in columns (b)-(d), and the numbers are normalised by the total number of edges $m$ of the corresponding dataset. We see that BN-FS-MARS (BFM) almost always has the highest number of TPs as well as the highest sums of TP and WD. GES reports the second highest numbers, and BIC reports the lowest numbers, which are close to 0. This is in line with what we observed in F1-scores, showing that networks learned by BN-FS-MARS have a higher portion of edges present in the ground truth network, though some edges may not have the correct orientation. Note that FNs would correspond to the complement of the TP+WD plot, as they would show the number of edges that were not discovered at all.

Figure \ref{fig:Tetrad} (d) shows numbers for FP. GES almost always has the highest number of FPs, indicating a tendency to learn extra edges not present in the ground truth. BN-FS-MARS has the second highest FP, but these are usually significantly lower than those for GES. We observe that increasing sample size from $N=1000$ to $N=5000$ reduces the false positives significantly for both these methods. BIC2 has close to zero FPs. However, we can see that BIC2 had also failed to learn most of the edges in the network. 

 Figure \ref{fig:bnlearn} shows SHD results for bnlearn networks. It reports the SHD of the networks learned by each of the three methods in columns (b)-(d), and the numbers are normalised by the total number of edges $m$, of the corresponding dataset. Here we see that GES almost always has the highest number of TPs as well as the highest sums of TP and WD. BN-FS-MARS reports the second highest numbers, which are close to the GES numbers, and BIC2 reports the lowest numbers, which are often close to 0. 

For both Tetrad and bnlearn, we observe that increasing sample size from $N=1000$ to $N=5000$ usually improves TP and WD of networks learned using BN-FS-MARS and GES, and BN-FS-MARS is able to recover almost all of the edges in one of the instances (pigs). However, adding more data does not improve results for BIC2. For bnlearn benchmarks exceeding 1003 nodes, i.e. some of the munin sub-networks GES and BIC2 did not converge at the end of the time limit. GES stopped with an out of time error, and BIC2 ran out of the memory.

Figure \ref{fig:bnlearn} (d) shows numbers for FP for bnlearn. As with Tetrad networks, GES almost always has the highest number of FPs, indicating a tendency to learn extra edges not present in the ground truth. BN-FS-MARS has the second highest FP, but these are usually significantly lower than those for GES. We observe that increasing sample size from $N=1000$ to $N=5000$ sometimes leads to an increase in FPs for GES, but almost always a reduction for BN-FS-MARS. BIC2 has the close to zero FPs. However, as with Tetrad, BIC2 also failed to recover most ground truth edges. 

We note that for one of the benchmarks (diabetes), BN-FS-MARS has decreased F1-scores and TPs with $N=5000$. A limitation of using the MARS method on discrete values is that it attempts to optimise the set of basis functions for all models for the child variable. If the child's domain is of large size, the results for the overall aggregated model will suffer in performance. Random forests are also known to perform poorly for such variables. The benchmark diabetes has several variables with domains of size 20+, leading to the observed degradation of performance of BN-FS-MARS. The more erratic nature of the plots for bnlearn is also because of the very different ranges of variable domain sizes across the benchmarks.


\subsection{Performance on Continuous Variables}
 
 To test the performance of Algorithm \ref{alg:BNMARS} on continuous data, we used the 5 Dream4 networks, each of which have 100 nodes and between $176-249$ edges. We generated datasets using simulated steady-state measurements with the genenetweaver software, with 1000 and 5000 samples. We scored candidate parent sets of all nodes with BN-FS-MARS with random forests as the feature selection method. For scoring, we used the top 5 scoring candidate parent sets in Step 3 of Algorithm \ref{alg:BNMARS}. 
 Once the scoring was complete, we computed the DAG structures using GOBNILP (Step 11). Due to the anytime nature of the algorithm, we are able to obtain the best network found at any given time, and we use this property to report results on the network found within the time limit of 24 hours for score-and-search. We compare the performance of BN-FS-MARS with tabular BIC (using pyGOBNILP because of the ability of continuous data processing) and BIC based GES. 
 
\textbf{Performance on F1-scores:} Figure  \ref{fig:dream} reports the F1-score of networks learned by each of the three methods in column (a). For sample size $N=1000$, the methods have similar scores. Increasing sample size to $N=5000$ improves F1-scores of all methods slightly, and BN-FS-MARS shows the best scores. 

\textbf{Performance on SHD:} Figure \ref{fig:dream} shows SHD results for the networks learned by each of the three methods in columns (b)-(d), and the numbers are normalised by the total number of edges $m$ of the corresponding dataset. GES almost always has the highest number of TPs, but it is close to BN-FS-MARS in sums of TP and WD. BIC2 reports the lowest numbers for both of these cases, but unlike with the discrete benchmarks, the numbers are not close to zero.
For both Tetrad and bnlearn increasing sample size to $N=5000$ improves TP and WD of networks learned by all methods, and BN-FS-MARS has the highest numbers for TP+WD. 

The final column (d) shows numbers for FP. As with Tetrad and bnlearn networks, GES almost always has the highest number of FPs, indicating a tendency to learn extra edges not present in the ground truth. BN-FS-MARS has the second highest FP. We observe that increasing sample size to $N=5000$ sometimes leads to an increase in FPs for GES, but always a reduction for BN-FS-MARS. BIC2 has the lowest number of FPs.
A change in the behaviour of these three methods can be because of the smaller network size ($n=100$). With a much smaller search space, BIC2 gives competitive results for F1-scores, in part because of its ability to learn fewer FPs. However, the strength of BN-FS-MARS comes from its feature selection mechanism, which is harder to solve for continuous data \citep{guyon2003introduction}.

\subsection{Model Complexity}

One of the strengths of BN-FS-MARS is its ability to identify large parent sets, as it does not pay the penalty of an exponential number of terms like  tabular BIC. 
Consider a variable in the benchmark pathfinder with a domain of size 4. It has five parents of domains sizes between $2-3$. Tabular BIC would need almost 300 parameters to represent it in CPT form, and it would give it a large penalty, making it difficult to be able to detect it with score-and-search. BN-FS-MARS assigns it a model with 32 terms, with none of the basis functions having a product of more than two terms. The lowered penalty makes such a parent set more likely to be selected, especially if it has a good score. In our experiments we found that modeling parent sets with BN-FS-MARS enabled us to score very large parent sets easily.

\section{Conclusion}\label{SECTION:conclusion}

In this paper, we propose a novel approach to score-and-search  for learning Bayesian networks with MARS relations as possible representations for CPDs. This algorithm lets us model non-linear relationships between nodes with low complexity, enabling us to learn large parent set sizes, and it does not place any constraints on the global network structure. Our approach can effectively prune most candidate parent sets of a variable by leveraging variable importance results from feature learning algorithms. We show empirically that this algorithm can solve both discrete and continuous instances with a very large number of nodes. 
\vskip 0.2in
\bibliography{scip,probabilistic}
\end{document}